\documentclass[10pt, a4paper]{article}
\usepackage{lrec}
\usepackage{mcite}
\usepackage{multibib}
\newcites{languageresource}{Language Resources}
\usepackage{graphicx}
\usepackage{tabularx}
\usepackage{soul}

\usepackage{epstopdf}
\usepackage[utf8]{inputenc}

\usepackage{hyperref}
\usepackage{xstring}

\usepackage{color}

\newcommand{\dpseg}{\texttt{dpseg}}

\title{Investigating Language Impact in Bilingual Approaches for \\Computational Language Documentation}

\name{Marcely Zanon Boito$^{1}$, Aline Villavicencio$^{2,3}$, Laurent Besacier$^{1}$}

\address{ (1) Laboratoire d'Informatique de Grenoble (LIG), UGA, G-INP, CNRS, INRIA, France\\ 
    (2) Department of Computer Science, University of Sheffield, England \\ 
    (3) Institute of Informatics (INF), UFRGS, Brazil \\ 
         \textbf{contact:} marcely.zanon-boito@univ-grenoble-alpes.fr \\}

\abstract{
For endangered languages, data collection campaigns have to accommodate the challenge that many of them are from oral tradition, and producing transcriptions is costly.
Therefore, it is fundamental to  translate them into a widely spoken language to ensure interpretability of the recordings. In this paper we investigate how the choice of translation language affects the posterior documentation work and potential automatic approaches which will work on top of the produced bilingual corpus. 
For answering this question, we use the MaSS multilingual speech corpus~\cite{boito2019mass} for creating 56 bilingual pairs that we apply to the task of low-resource unsupervised word segmentation and alignment. 
Our results highlight that the choice of language for translation influences the word segmentation performance, and that different lexicons are learned by using different aligned translations. 
Lastly, this paper proposes a \textit{hybrid} approach for bilingual word segmentation, combining \textit{boundary clues} extracted from a non-parametric Bayesian model~\cite{goldwater2009bayesian} with the attentional word segmentation neural model from \newcite{godard2018unsupervised}. Our results suggest that 
incorporating these clues into the neural models' input representation increases their translation and alignment quality, specially for challenging language pairs. \\ \newline 
}

\begin{document}

\maketitleabstract
\section{Introduction}
Computational Language Documentation~(CLD) is an emerging research field whose focus lies on helping to automate the manual steps performed by linguists during language documentation. The need for this support is ever more crucial given predictions that more than 50\% of all currently spoken languages will vanish before 2100~\cite{austin2011cambridge}. For these very low-resource scenarios, transcription is very time-consuming: one minute of audio is estimated to take one hour and a half on average of a linguist's work~\cite{austin2013endangered}. 

This \textit{transcription bottleneck} problem~\cite{brinckmann2009transcription}, combined with a lack of human resources and time for documenting all these endangered languages, can be attenuated by translating into a widely spoken language to ensure subsequent interpretability of the collected recordings. 
Such parallel corpora have been recently created by aligning the collected audio with translations in a well-resourced language~\mcite{adda2016breaking, godard2017very, boito2018small}, and some linguists even suggested that more than one translation should be collected to capture deeper layers of meaning~\cite{evans2004searching}.
However, in this documentation scenario, the impact of the language chosen for translation rests understudied, and it is unclear if similarities among languages have a significant impact in the automatic bilingual methods used for information extraction (these include word segmentation, word alignment, and translation).

Recent work in CLD includes the use of aligned translation for improving transcription quality~\cite{anastasopoulos2018leveraging}, and for obtaining bilingual-rooted word segmentation~\cite{duong2016attentional,boito2017unwritten}. There are pipelines for obtaining manual~\cite{foley2018building} and automatic~\cite{michaud2018integrating} transcriptions, and for aligning transcription and audio~\cite{strunk2014untrained}. Other examples are methods for low-resource segmentation~\cite{lignos2010recession, Goldwater09bayesian}, and for lexical unit discovery without textual resources~\cite{bartels2016toward}. Moreover, direct speech-to-speech~\cite{tjandra2019speech} and speech-to-text~\cite{besacier2006towards, berard2016listen} architectures could be an option for the lack of transcription, but there is no investigation yet about how exploitable these architectures can be in low-resource settings. Finally, previous work also showed that Neural Machine Translation models at the textual level are able to provide exploitable soft-alignments between sentences by using only 5,130 training examples ~\cite{boito2019empirical}.

In this work, we investigate the existence of language impact in bilingual approaches for CLD, tackling word segmentation,\footnote{Here, word is defined as a sequence of phones that build a minimal unit of meaning.} one of the first tasks performed by linguists after data collection.
More precisely, the task consists in detecting word boundaries in an unsegmented phoneme sequence in the language to document, supported by the translation available at the sentence-level. The phonemes in the language to document can be manually obtained, or produced automatically as in \newcite{godard2018unsupervised}.

For our experiments, we use the eight languages from the multilingual speech-to-speech MaSS dataset~\cite{boito2019mass}: Basque~(EU), English~(EN), Finnish~(FI), French~(FR), Hungarian~(HU), Romanian~(RO), Russian~(RU) and Spanish~(ES).
We create 56 bilingual models, seven per language, simulating the documentation of each language supported by different sentence-level aligned translations. This setup allows us to investigate how having the same content, but translated in different languages, affects bilingual word segmentation.  
We highlight that in this work we use a dataset of well-resourced languages due to the lack of multilingual resources in documentation languages that could be used to investigate this hypothesis. Thus, for keeping results coherent and generalizable for CLD, we down-sample our corpus, running our experiments using only 5k aligned sentences as a way to simulate a low-resource setting.
We train bilingual models based on the segmentation and alignment method from \newcite{godard2018unsupervised}, investigating the language-related impact in the quality of segmentation, translation and alignment.

Our results confirm that the language chosen for translation has a significant impact on word segmentation performance, what aligns with \newcite{haspelmath2011indeterminacy} who suggests that the notion of word cannot always be meaningfully defined cross-linguistically. We also verify that joint segmentation and alignment is not equally challenging across different languages: 
while we obtain good results for EN, the same method fails to segment the language-isolate EU. 
Moreover, we verify that the bilingual models trained with different aligned translations learn to focus on different structures, what suggests that having more than one translation could enrich computational approaches for language documentation.
Lastly, the models' performance is 
improved by the introduction of a \textit{hybrid} approach, which leverages the \textit{boundary clues} obtained by a monolingual non-parametric Bayesian model~\cite{Goldwater09bayesian} into the bilingual models. This type of intermediate annotation is often produced by linguists during documentation, and its incorporation into the neural model can be seen as a form of validating word-hypotheses. 

This paper is organized as follows. Section~\ref{sec2} presents the models investigated for performing word segmentation. Section~\ref{sec3} presents the experimental settings, and Section~\ref{sec4} the results and discussion. 
Section~\ref{sec5} concludes the work.

\section{Models for Word Segmentation} \label{sec2}

\begin{figure*}
\centering
\includegraphics[scale=0.59]{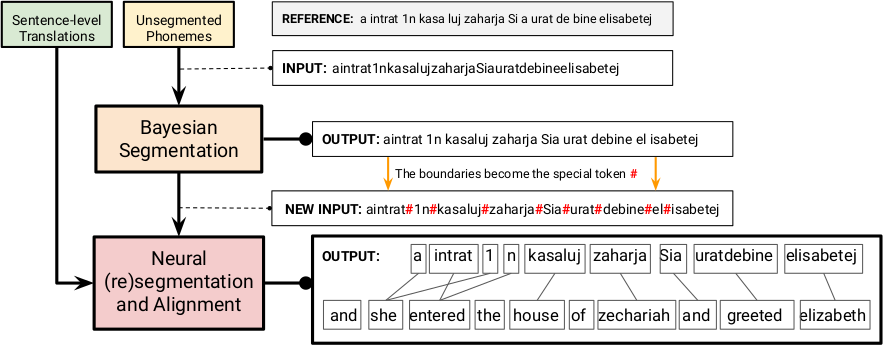}
      \caption{An illustration of the hybrid pipeline for the EN$>$RO language pair. The Bayesian model receives the unsegmented phonemes, outputing segmentation. The discovered boundaries are then replaced by a special token, and bilingual re-segmentation and alignment are jointly performed.}
      \label{fig:example}
\end{figure*}

\subsection{Monolingual Bayesian Approach}\label{met:mono}
Non-parametric Bayesian models~\cite{goldwater2007nonparametric,johnson2009improving} are statistical approaches that can be used for word segmentation and morphological analysis, being known as very robust in low-resource settings~\cite{godard2016preliminary,goldwater2009bayesian}. In these  
monolingual models, words are generated by a uni or bigram model over a non-finite inventory, through the use of a Dirichlet process.
Although providing reliable segmentation in low-resource settings, these monolingual models are incapable of automatically producing alignments with a foreign language, and therefore the discovered pseudo-word segments can be seen as ``meaningless''. \newcite{godard2018unsupervised} also showed that \dpseg{}\footnote{Available at \url{http://homepages.inf.ed.ac.uk/sgwater/resources.html}}  \cite{Goldwater06contextual,goldwater2009bayesian} behaves poorly on pseudo-phone units discovered from speech, which limits its application. 
Here, we investigate its use as an intermediate monolingual-rooted segmentation system, whose discovered boundaries are used as clues by bilingual models. 

\subsection{Bilingual Attention-based Approach}\label{met:neural}
We reproduce the approach from \newcite{godard2018unsupervised} who train Neural Machine Translation~(NMT) models between language pairs, using as source language the translation (word-level) and as target, the language to document (unsegmented phoneme sequence). 
Due to the attention mechanism present in these networks~\cite{bahdanau2014neural}, posterior to training, it is possible to retrieve \textit{soft-alignment probability matrices} between source and target sentences.

The soft-alignment probability matrix for a given sentence pair is a collection of context vectors. Formally, a context vector for a decoder step $t$ is computed using the set of source annotations $H$ and the last state of the decoder network ($s_{t-1}$, the translation context). The attention is the result of the weighted sum of the source annotations $H$ (with $H = h_1,..., h_A$) and their $\alpha$ probabilities (Eq. \ref{seq2seq-attention}). Finally, these are obtained through a feed-forward network $align$, jointly trained, and followed by a softmax operation (Eq. \ref{seq2seq-attention2}).

\begin{equation}\label{seq2seq-attention}
c_t = 
Att(H, s_{t-1})= \sum_{i=1}^{A}\alpha_i^t h_i
\end{equation}
\begin{equation}\label{seq2seq-attention2}
\alpha_i^t = {\rm softmax}(align(h_i, s_{t-1})) 
\end{equation}

The authors show that these soft-alignment probability matrices can be used to produce segmentation over phoneme (or grapheme) sequences. This is done by segmenting neighbor phonemes whose probability distribution (over the words in the aligned source translation) peaks at different words. The result is a pair of phoneme sequences and translation words, as illustrated on the bottom half of Figure~\ref{fig:example}. 
In this work we refer to this type of model simply as \textbf{neural model}.

\subsection{Bilingual Hybrid Approach}\label{met:hybrid}

The monolingual approach~($\S$\ref{met:mono}) has the disadvantage of not producing bilingual alignment, but it segments better than the bilingual approach~($\S$\ref{met:neural}) when the phonemic input is used~\cite{godard2018unsupervised}. In this work we investigate a simple way of combining both approaches by creating a \textit{hybrid} model which takes advantage of the Bayesian method's ability to correctly segment from small data while jointly producing translation alignments.

We augment the original unsegmented phoneme sequence with the \dpseg{} output boundaries. 
In this augmented input representation, illustrated in Figure~\ref{fig:example}, a boundary is denoted by a special token which separates the words identified by \dpseg{}. We call this \textit{soft-boundary insertion}, since the \dpseg{} boundaries inserted into the phoneme sequence can be ignored by the NMT model, and new boundaries can be inserted as well. For instance, in Figure~\ref{fig:example} \texttt{aintrat} becomes \texttt{a intrat} (boundary insertion), and \texttt{urat debine} becomes \texttt{uratdebine} (soft-boundary removal). 

\section{Experimental Settings}\label{sec3}
\noindent\textbf{Multilingual Dataset:} For our experiments we use the MaSS dataset~\cite{boito2019mass}, a 
fully aligned and multilingual dataset containing 8,130 sentences extracted from the Bible. 
The dataset provides multilingual speech and text alignment between all the available languages: English~(EN), Spanish~(ES), Basque~(EU), Finnish~(FI), French~(FR), Hungarian~(HU), Romanian~(RO), Russian~(RU). As sentences in documentation settings tend to be short, we used RO as the pivot language for removing sentences longer (in terms of number of tokens) than 100 symbols. The resulting corpus contains 5,324 sentences, a size which is compatible with real language documentation scenarios. Table~\ref{tab:corpusinfo} presents some statistics. 
For the phonemic transcription of the speech (target side of the bilingual segmentation pipeline), we use the automatic phonemization from \textit{Maus forced aligner}~\cite{kisler2017multilingual}, which results in an average vocabulary reduction of 835 types, the smallest being for RO (396), and the most expressive being for FR (1,708). This difference depends on the distance between phonemic and graphemic forms for each language. The phonemizations present an average number of unique phonemes of 42.5.
Table~\ref{tab:corpusinfo2} presents the statistic for the phonemic representation.

\noindent\textbf{Training and Evaluation:} 
For monolingual segmentation, we use \dpseg{}'s unigram model with the same hyperparameters as \newcite{godard2016preliminary}. The bilingual neural models were trained using a one-layer encoder (embeddings of 64), and a two-layers decoder (embeddings of 16). The remaining parameters come from \newcite{godard2018unsupervised}. From this work, we also reproduced the \textit{multiple runs averaging}: for every language pair, we trained two networks, averaging the soft-alignment probability matrices produced. This averaging can be seen as \textit{agreement} between the alignment learned with different parameters initialization.
Regarding the 
data, 10\% of the multilingual ids were randomly selected for validation, and the remaining were used for training. We report BLEU scores~\cite{papineni2002bleu} over the validation set for assessing translation quality. For hybrid setups, the soft-boundary special token is removed from the output before scoring, so results are comparable.
Finally, for the reported word discovery results, the totality of the corpus is considered for evaluation.

\begin{table}
\centering
\begin{tabular}{lcccc}
\hline
            & \multicolumn{1}{l}{\textbf{\#Types}} & \multicolumn{1}{l}{\textbf{\#Tokens}} & \multicolumn{1}{l}{\textbf{\begin{tabular}[c]{@{}l@{}}Token \\ Length\end{tabular}}} & \multicolumn{1}{l}{\textbf{\begin{tabular}[c]{@{}l@{}}Token/\\ Sentence\end{tabular}}} \\\hline
\textbf{EN} & 5,232                                & 90,716                                & 3.98                                                                                 & 17.04                                                                                      \\
\textbf{ES} & 8,766                                & 85,724                                & 4.37                                                                                 & 16.10                                                                                      \\
\textbf{EU} & 11,048                               & 67,012                                & 5.91                                                                                 & 12.59                                                                                      \\
\textbf{FI} & 12,605                               & 70,226                                & 5.94                                                                                 & 13.19                                                                                      \\
\textbf{FR} & 7,226                                & 94,527                                & 4.12                                                                                 & 17.75                                                                                      \\
\textbf{HU} & 13,770                               & 69,755                                & 5.37                                                                                 & 13.10                                                                                      \\
\textbf{RO} & 7,191                                & 88,512                                & 4.06                                                                                 & 16.63                                                                                      \\
\textbf{RU} & 11,448                               & 67,233                                & 4.66                                                                                 & 12.63     \\ \hline                                                                                
\end{tabular}
\caption{Statistics for the textual portion of the corpus. The last two columns bring the average of the named metrics.}
\label{tab:corpusinfo}
\end{table}

\begin{table}
\centering
\begin{tabular}{lcccc}
\hline
            & \textbf{\#Types} & \multicolumn{1}{l}{\textbf{\#Tokens}} & \textbf{\begin{tabular}[c]{@{}c@{}}Token \\ Length\end{tabular}} & \textbf{\begin{tabular}[c]{@{}c@{}}Phonemes/\\ Sentence\end{tabular}} \\\hline
\textbf{EN} & 4,730            & 90,657                                & 3.86                                                             & 56.18                                                                 \\
\textbf{ES} & 7,980            & 85,724                                & 4.30                                                             & 68.52                                                                 \\
\textbf{EU} & 9,880            & 67,012                                & 6.94                                                             & 71.13                                                                 \\
\textbf{FI} & 12,088           & 70,226                                & 5.97                                                             & 72.37                                                                 \\
\textbf{FR} & 5,518            & 93,038                                & 3.21                                                             & 52.86                                                                 \\
\textbf{HU} & 12,993           & 69,755                                & 5.86                                                             & 65.52                                                                 \\
\textbf{RO} & 6,795            & 84,613                                & 4.50                                                             & 68.04                                                                 \\
\textbf{RU} & 10,624           & 67,176                                & 6.19                                                             & 59.26              \\\hline                                                  
\end{tabular}
\caption{Statistics for the phonemic portion of the corpus. The last two columns bring the average of the named metrics.}
\label{tab:corpusinfo2}
\end{table}

\section{Bilingual Experiments}\label{sec4}

Word segmentation boundary F-scores are presented in Table~\ref{tab:bilingualresults}. 
For the bilingual methods, Table~\ref{tab:bleuscores} presents the averaged BLEU scores. We observe that, similar to the trend observed in Table~\ref{tab:bilingualresults}, hybrid models are in average superior in terms of BLEU scores.\footnote{We find an average BLEU scores difference between best hybrid and neural setups of 1.50 points after removing the outlier (RO). For this particular case, hybrid setups have inferior translation performance (average BLEU reduction of 11.44).} Moreover, we observe that segmentation and translation scores are strongly correlated for six of the eight languages, with an average $\rho$-value of $0.76$ (significant to $p<0.01$). The exceptions were EU~(0.46) and RO~(-0.06). While for EU we believe the general lack of performance of the systems could explain the results, the profile of RO hybrid setups was surprising. It highlights that the relationship between BLEU score and segmentation performance is not always clearly observed. In summary, we find that the addition of soft-boundaries will increase word segmentation results, but its impact to translation performance needs further investigation.

Looking at the segmentation results, we verify that, given the same amount of data and supervision, the segmentation performance for different target languages vary: EN seems to be the easiest to segment (neural:~69.1, hybrid:~73.3), while EU is the most challenging to segment with the neural approach (neural:~38.4, hybrid:~47.3).  The following subsections will explore the relationship between segmentation, alignment performance and linguistic properties.

\begin{table}
\centering
\includegraphics[scale=0.44]{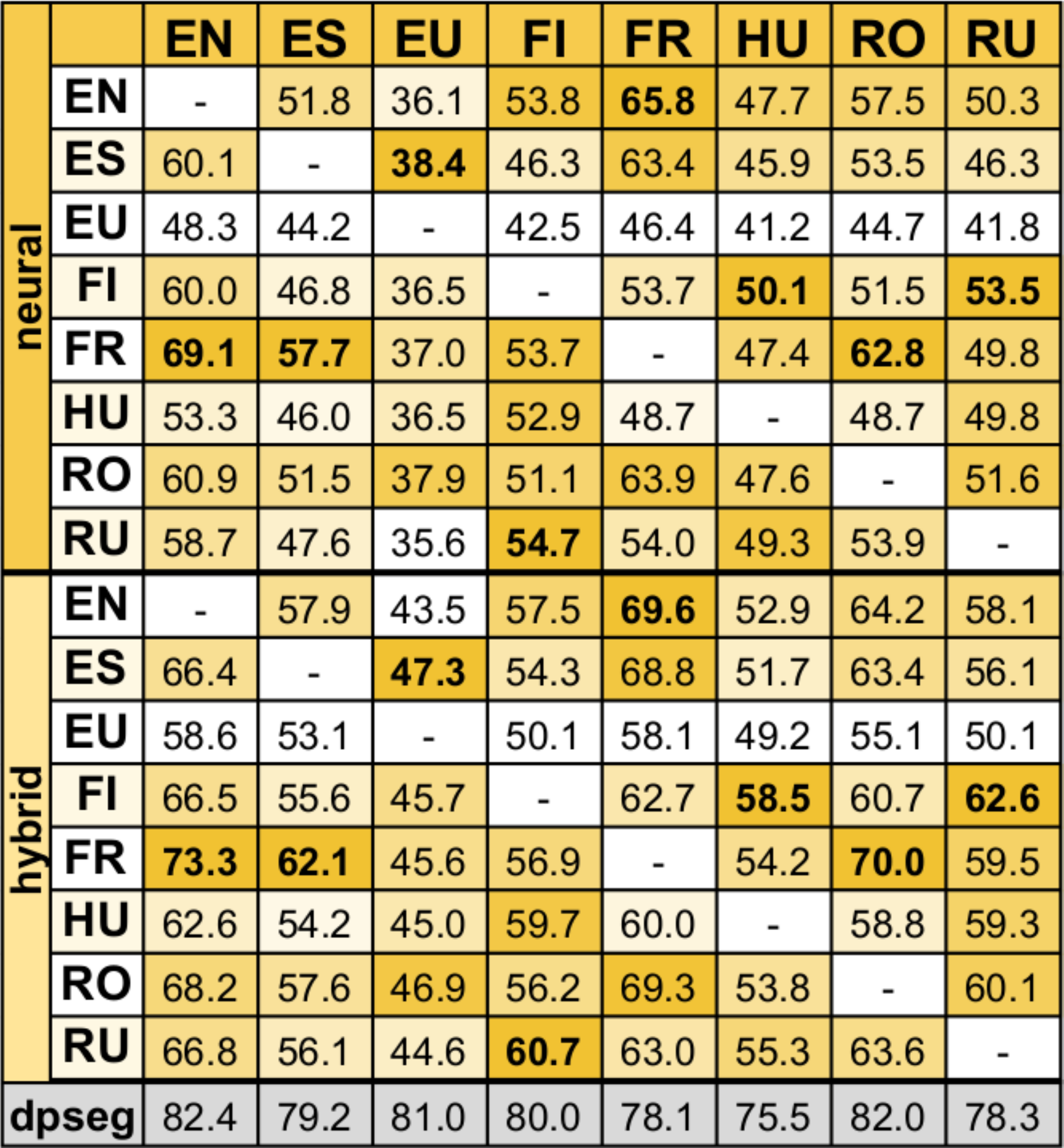}
\caption{Word Segmentation Boundary F-score results for neural (top), hybrid (middle) and \dpseg~(bottom). The columns represent the target of the segmentation, while the rows represented the translation language used. For bilingual models, darker squares represent higher scores.
Better visualized in color.}
\label{tab:bilingualresults}
\end{table}
\begin{table}
\centering
\includegraphics[scale=0.44]{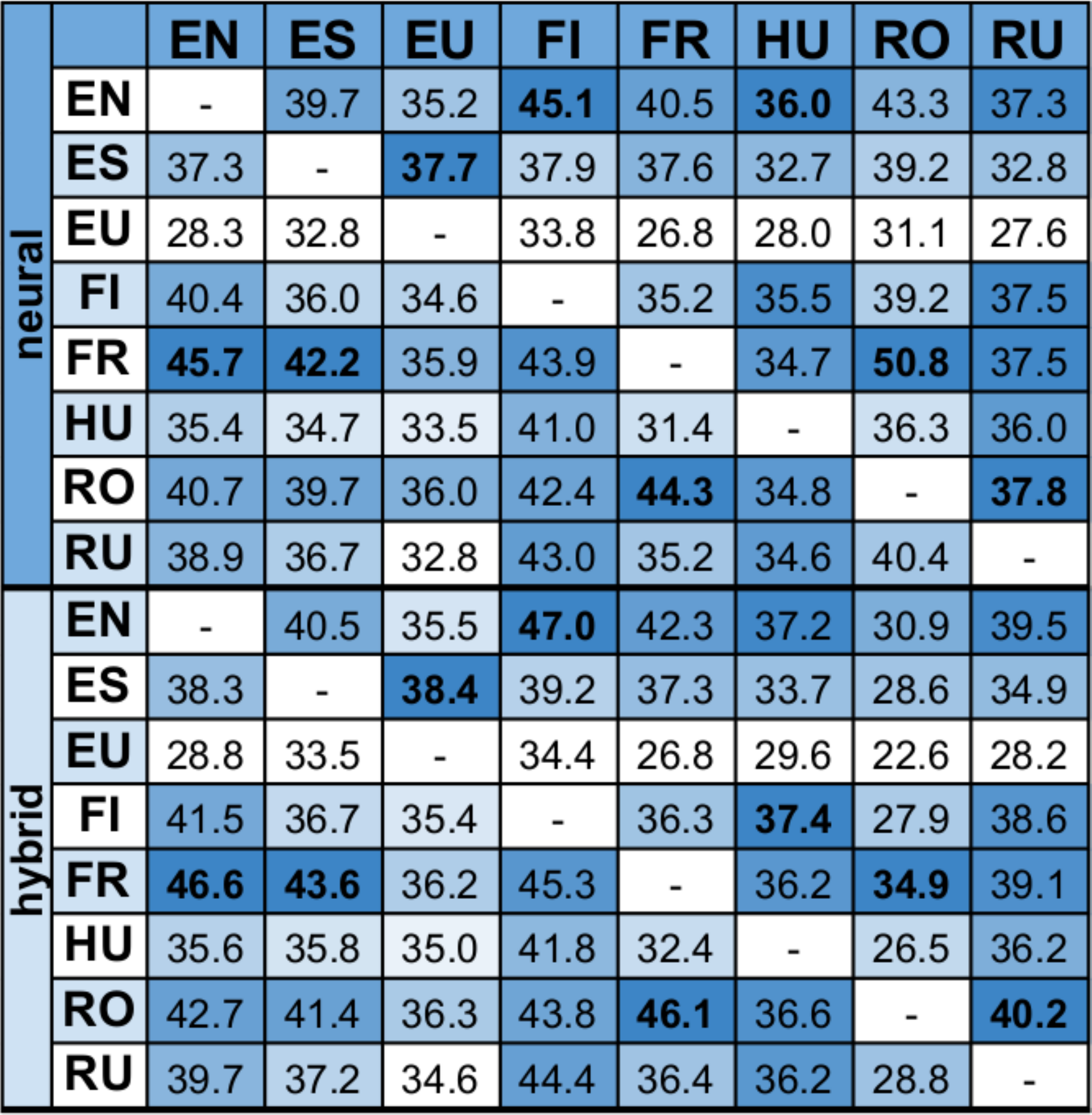}
\caption{BLEU 4 average results for neural~(top) and hybrid~(bottom) bilingual models. The columns represent the target of the segmentation. Darker squares represent higher scores. Better visualized in color.}
\label{tab:bleuscores}
\end{table}

\subsection{Source Language Impact}

\paragraph{Bilingual Baseline Comparison:}
The results confirm that there is an impact related to using different source languages for generating the segmentations, and we identify interesting language pairs emerging as the most efficient, such as FI$>$HU (Uralic Family), FR$>$RO and FR$>$ES (Romance family).\footnote{We denote L1$>$L2 as using L1 for segmenting L2. L1$<>$L2 means L1$>$L2 and L2$>$L1.} 
In order to consolidate these results, we investigate if the  language ranking obtained (in terms of \textit{best translation languages for segmenting a target language}) is due to a similar profile of the source and target languages in terms of word length and tokens per sentence. Since translation words are used to cluster the phoneme sequences into words (bilingual-rooted word segmentation), having more or less translation words could be a determining aspect in the bilingual segmentation performed (more details about this in Section~\ref{sec:vocabstudy}).
For this investigation, we use a naive bilingual baseline called proportional~\cite{godard2018unsupervised}. It performs segmentation by distributing phonemes equally between the words of the aligned translation, insuring that words that have more letters, receive more phonemes (hence \textit{proportional}). The average difference between the best hybrid~(Table~\ref{tab:bilingualresults}) and proportional~(Table~\ref{tab:proportional}) results is of 25.92 points.
This highlights not only the challenge of the task, but that the alignments learned by the bilingual models are not trivial.

\begin{table}
\centering
\includegraphics[scale=0.46]{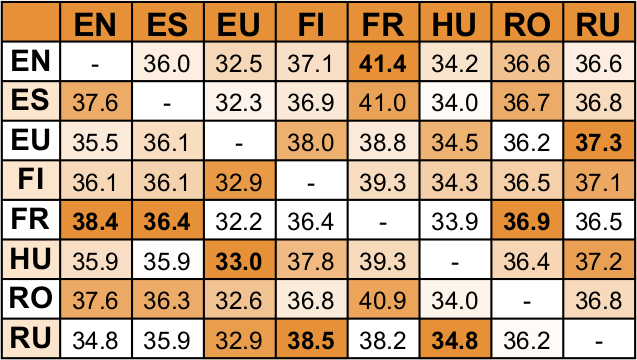}
      \caption{Proportional segmentation results. The columns represent the target of the segmentation. Darker squares represent higher word boundary F-scores. Better visualized in color.}
      \label{tab:proportional}
\end{table}

We compute Pearson's correlation between bilingual hybrid and proportional 
segmentation scores, observing that no language presents a significant correlation for $p<0.01$. However, when all languages pairs are considered together ($N=56$), a significant positive correlation (0.74) is observed. Our interpretation is that the token ratio between the number of tokens in source and the number of tokens in target sentences have a significant impact on bilingual segmentation difficulty. However, it does not, by itself, dictates the best choice of translation language for a documentation scenario. For instance, the proportional baseline results indicate that EU is the best choice for segmenting RU. This choice is not only linguistically incoherent, but bilingual models reached their worst segmentation and translation results by using this language. This highlights that while statistical features might impact greatly low-resource alignment and should be taken into account, relying only on them might result in sub-optimal models.

\paragraph{Language Ranking:}
Looking into 
the quality of the segmentation results and their relationship with the language ranking, our intuition was that languages from the same family would perform the best. For instance, we expected ES$<>$FR, ES$<>$RO, FR$<>$RO (Romance family) and FI$<>$HU (Uralic family) to be strong language pairs. 
While some results confirm this hypothesis (FR$>$ES, FI$>$HU, FR$>$RO), the exceptions are: EN$>$FR, RU$<>$FI and ES$>$EU. 
For EN$>$FR, we argue that EN was ranked high for almost all languages, which could be due to some convenient statistic features. Table~\ref{tab:corpusinfo} shows that EN presents a very reduced vocabulary in comparison to the other languages. This could result in an easier language modeling scenario, which could then reflect in a better alignment capacity of the trained model. Moreover, for this and for RU$<>$FI scenarios, results seemed to reproduce the trend from 
the proportional baseline, in which these pairs were also found to be the best. This could be the result of a low syntactic divergence between languages of these pairs. 
Finally, the language isolate EU is not a good choice for segmenting any language (worst result for all languages). If we consider that this language has no relation to any other in this dataset, this result could be an indication that documentation should favor languages somehow related to the language they are trying to document. In fact, results for EU segmentation are both low (F-score and BLEU) and very close to the proportional baseline (average difference of 4.23 for neural and 13.10 for hybrid), which suggests  that these models were not able to learn meaningful bilingual alignment.

\subsection{Hybrid Setups}
Looking at the hybrid results, we verify that the these models outperform their neural counterparts. 
Moreover, the impact of having the \textit{soft-boundaries} is larger for the languages whose bilingual segmentation seems to be more challenging, hinting that the network is learning to leverage the \textit{soft-boundaries} for generating a better-quality alignment between challenging language pairs. Table~\ref{tab:vocabdpseg} presents the intersection between the correct types discovered by both monolingual and hybrid models. Results show that while the monolingual baseline \textit{informs} the bilingual models, it is not completely responsible for the increase in performance. 
This hints that giving boundary clues to the network will not simply force some pre-established segmentation, but instead it will enrich the network's internal representation.
Moreover, it is interesting to observe that the degree of overlap between the vocabulary generated will depend on the language target of segmentation, hinting that some languages might \textit{accept} more easily the \textit{soft-boundaries} proposed by the monolingual approach.

\begin{table}
\centering
\includegraphics[scale=0.37]{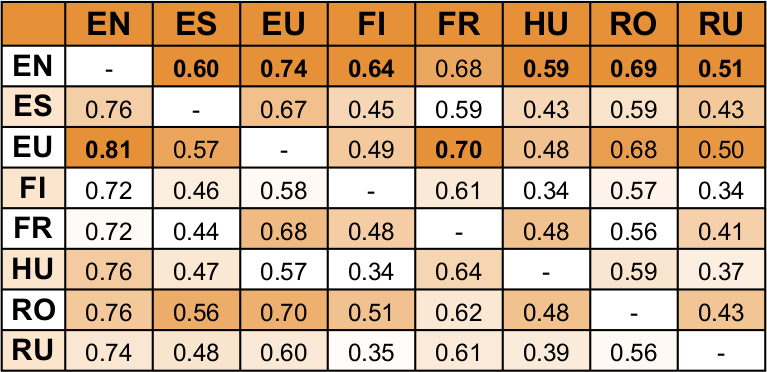}
      \caption{Intersection between the correct types discovered by both monolingual and hybrid models. We notice that the target language of the segmentation (columns) has an impact in the acceptance of soft-boundaries by the neural model.}
      \label{tab:vocabdpseg}
\end{table}

Nonetheless, compared to the monolingual segmentation~(Table~\ref{tab:bilingualresults}), even if the hybrid approach improves over the base neural one, it deteriorates considerably the performance with respect to \dpseg{} (average difference of 16.54 points between the best hybrid result and its equivalent monolingual segmentation). However, this deterioration is necessary in order to discover
semantically meaningful structures (joint bilingual segmentation and alignment), which is a harder task than monolingual segmentation.
In this scenario, the monolingual results should be interpreted as an intermediate, good quality, segmentation/word-hypotheses created by linguists, which might be validated or not in light of the system's bilingual output. 

\subsection{Analysis of the Discovered Vocabulary}\label{sec:vocabstudy}
Next we study the characteristics of the vocabulary outputed by the bilingual models focusing on the impact caused by the aligned translation. For this investigation, we report results for hybrid models only, since their neural equivalents present the same trend. We refer as \textit{token} the collection of phonemes segmented into word-like units. \textit{Types} are defined as the set of distinct tokens. Table~\ref{tab:tokenlength} brings the hybrid model's total number of types.

Looking at the rows, we see that EN, ES, FR, RO, which are all fusional languages, generated in average the smallest vocabularies. We also notice that HU and FI are the languages that tend to create the largest vocabularies when used as translation language. This could be due to both languages accepting a flexible word order, thus creating a difficult alignment scenario for low-resource settings. Moreover, these languages, together with EU, are agglutinative languages. This might be an explanation for the lack of performance in general for setups using these languages as target. In these conditions, the network must learn to align many translation words to the same structure in order to achieve the expected segmentation. However, sometimes  over-segmentation might be the result of the network favoring alignment content instead of phoneme clustering. 

\begin{figure}
\centering
\includegraphics[scale=0.52]{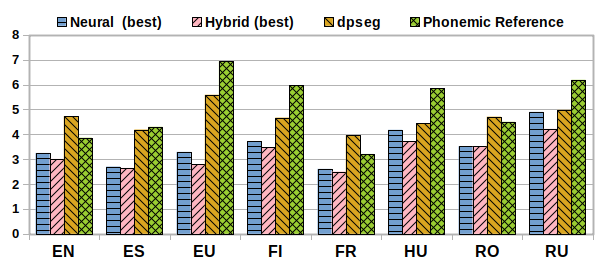}
\caption{Average token length of the reference, monolingual \dpseg , and best neural and hybrid setups from Table~\ref{tab:bilingualresults}.}
\label{tab:vocabulary}
\end{figure}

Notwithstanding, the models for agglutinative languages are not the only ones over-segmenting.
Looking at the average token length of the  segmentations produced in Figure~\ref{tab:vocabulary}, and supported by the size of the vocabularies, we verify that bilingual approaches tend to over-segment the output independent of the language targeted. This over-segmentation tends to be more accentuated in hybrid setups, with the exception of EN, FR and RO. This is probably due to the challenge of clustering the very long sequence of phonemes into the many available source words (see statistics for words and phonemes per sentence in Tables~\ref{tab:corpusinfo} and \ref{tab:corpusinfo2}).

\begin{figure}
\centering
\includegraphics[scale=0.59]{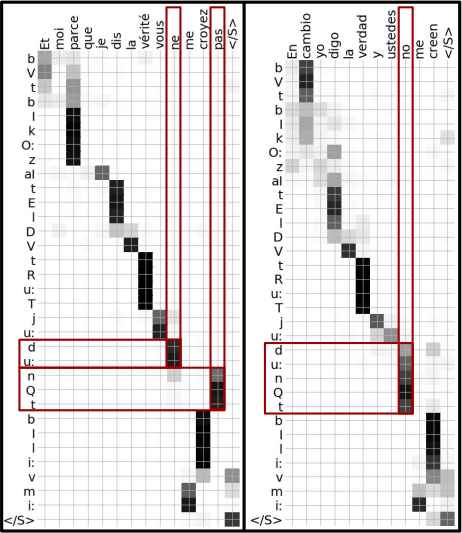}
      \caption{EN attention matrices generated by neural FR~(left) and ES~(right) bilingual models. The squares represent alignment probabilities (the darker the square, the higher the probability). The EN phonemization (rows) correspond to the following sentence: ``But because I tell the truth, you do not believe me''. }
      \label{fig:segexample}
\end{figure}

Furthermore, the very definition of a word might be difficult to define cross-linguistically, as discussed by \newcite{haspelmath2011indeterminacy}, and different languages might encourage a more fine-grained segmentation. 
For instance, in Figure~\ref{fig:segexample} we see the EN alignment generated by the FR and ES neural models for the same sentence. Focusing at the \textit{do not} (\texttt{du:nQt}) at the end of the sentence, we see that the ES model does not segment it, aligning everything to the ES translation \texttt{no}. Meanwhile the FR model segments the structure in order to align it to the translation \texttt{ne pas}. In both cases the discovered alignments are correct however, the ES segmentation is considered wrong. This highlights that the use of a segmentation task for evaluating the learned alignment might be sub-optimal, and that a more in-depth evaluation of source-to-target correspondences should be considered. In Section~\ref{sec:alignment} we showcase a method for filtering the alignments generated by the bilingual models. 

\begin{table}
\centering
\includegraphics[scale=0.37]{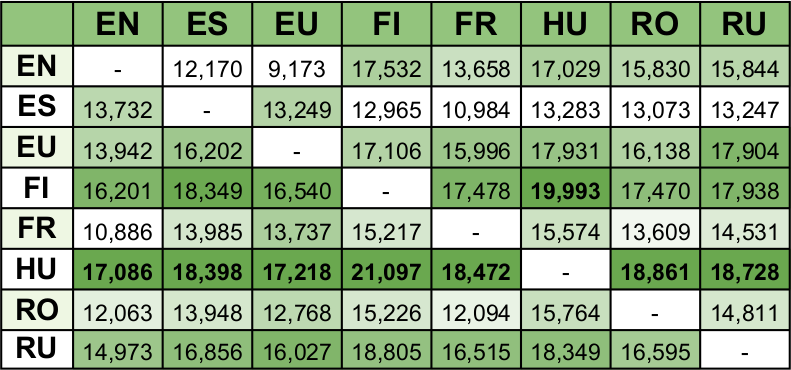}
\caption{Number of types produced by the hybrid models.}
\label{tab:tokenlength}
\end{table}

Concluding, in this work we study the alignment \textit{implicitly} optimized by a neural model. An interesting direction would be the investigation of explicit alignment optimization for translation models, such as performed in \newcite{godard2019controlling}, where the authors consider the segmentation length generated by the bilingual alignments as part of their loss during training.

\subsection{Alignment Confidence}\label{sec:alignment}

The neural approach used here for bilingual-rooted word segmentation produces alignments between source and target languages. In this section we investigate how these alignments vary in models trained using different translation (source) languages. This extends the results from the previous section, that showed that models trained on different languages will present different lexicon sizes. We aim to show that this difference in segmentation behavior comes from the different alignments that are discovered by the models with access to different languages.

We use the approach from \newcite{boito2019empirical} for extracting the alignments the bilingual models are more \textit{confident about}. For performing such a task, \textit{Average Normalized Entropy}, as defined in \newcite{boito2019empirical}, is computed for every (segmentation, aligned translation) pair. The scores are used for ranking the alignments in terms of confidence, with low-entropy scores representing the high-confidence automatically generated alignments. In previous work, we showed that this approach allow us to increase type retrieval scores by filtering the good from the bad quality alignments discovered. For this investigation, we chose to present results applied to the target language FR.

\begin{table*}
\centering
\includegraphics[scale=0.7]{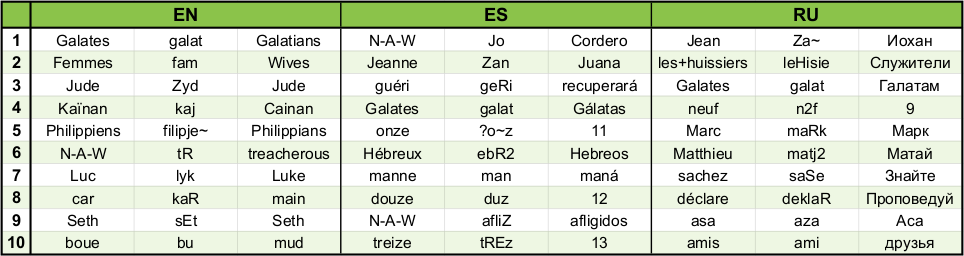}
      \caption{Top low-entropy/high-confidence (graphemization, phonemic segmentation, aligned translation) results for EN, ES and RU models for segmenting FR. The output of the system is the phonemic segmentation, and graphemization is provided only for readability purpose. 
      N-A-W identify unknown/incorrect generated types.}
      \label{tab:exfr}
\end{table*}

Table~\ref{tab:exfr} presents the top 10 low-entropy (high-confidence) pairs from 3 different translation languages
(from Table~\ref{tab:bilingualresults}, FR column). The phoneme sequences are accompanied by their grapheme equivalents to increase readability, but all presented results were computed over phoneme sequences. The other translation languages were also omitted for readability purpose. 

We observe a different set of discovered types depending on the language used, but all languages learn a fair amount of biblical names and numbers, very frequent due to the nature of the dataset.\footnote{The chapter names and numbers (e.g. ``Revelation 2'') are included in the dataset, totalizing 260 examples of \textit{``name, number''}.} 
This highlights that very frequent types might be captured independently of the language used, but other structures might be more dependent on the chosen language. We also notice the presence of incorrect alignments (the word \texttt{car} (because) aligned to the word \texttt{main}), concatenations (the words \texttt{les huissiers} (the ushers) became a single word) and incorrect types (\texttt{N-A-W} in the table). This is to be expected, as these are automatic alignments.

Confirming the intuition that the models are focused on different information depending on the language they are trained on, we studied the vocabulary intersection of the FR bilingual models for the top 200 correct discovered types ranked by alignment confidence.
We observed that the amount of shared lexicon for the sets is fairly small: the smallest intersection being of 20\%~(between EU and RO) and the largest one of 35.5\%~(between RU and FI). In other words, this means that the high-confidence alignments learned by distinct bilingual models differ considerably. Even for models that shared most structures,  
such as FI and RU~(35.5\%), and HU and RU~(34\%), this intersection is still limited. This shows that the bilingual models will discover different structures, depending on the supervision available. This is particularly interesting considering that the content of the aligned information remains the same, and the only difference between the bilingual models is the language in which the information is expressed. It highlights how collecting data in \textit{multilingual settings} (that is, in more than one translation language) could enrich approaches for CLD. Lastly, we leave as future work a more generalizable study of the distinctions in the bilingual alignments, including the evaluation of the word-level alignments discovered by the models.

\section{Conclusion} \label{sec5}

In language documentation scenarios, transcriptions are most of the time costly and difficult to obtain. In order to ensure the interpretability of the recordings, a popular solution is to replace manual transcriptions by translations of the recordings in well-resourced languages~\cite{adda2016breaking}. However, while some work suggests that translations in multiple languages may capture deeper layers of meaning~\cite{evans2004searching}, most of the produced corpora from documentation initiatives are bilingual. Also, there is a lack of discussion about the impact of the language chosen for these translations in posterior automatic methods.

In this paper we investigated the existence of language-dependent behavior in a bilingual method for unsupervised word segmentation, one of the first tasks performed in post-collection documentation settings. We simulated such a scenario by using the MaSS dataset~\cite{boito2019mass} for training 56 bilingual models, the combination of all the available languages in the dataset. 
Our results show that in very low-resource scenarios (only 5,324 aligned sentences), the impact of language can be great, with a large margin between best and worst results for every target language. We also verify that the languages are not all equally difficult to segment.
Moreover, while some of our \textit{language rankings}, in terms of best translation languages for segmenting a target language, could be explained by the linguistic family of the languages (FR$>$ES, FI$>$HU, FR$>$RO), we found some surprising results such as ES$>$EU and EN$>$FR. We believe these are mostly due to the impact of existing statistic features~(e.g. token length ratio between source and target sentences, and vocabulary size), related to the corpus, and not to the language features.

Additionally, we investigated providing a different form of supervision to the bilingual models. We used the monolingual-rooted segmentation generated by \dpseg{} for augmenting the phoneme sequence representation that the neural models learn from at training time. We observed that the networks learned to leverage \dpseg{}'s \textit{soft-boundaries} as hints of alignment break (boundary insertion). Nonetheless, the networks are still robust enough to ignore this information when necessary.
This suggests that, in a documentation scenario, \dpseg{} could be replaced by early annotations of potential words done by a linguist, for instance. The linguist could then validate the output of the neural system, and review their word hypotheses considering the generated bilingual alignment.

In summary, our results highlight the existence of a relationship between language features and performance in (neural) bilingual segmentation. We verify that languages close in phonology and linguistic family score better, while less similar languages yield lower scores. While we find that our results are rooted in linguistic features, we also believe there is a non-negligible relationship with corpus statistic features which can impact greatly neural approaches in low-resource settings.

\section{Bibliographical References}\label{reference}

\bibliographystyle{lrec}
\bibliography{emnlp-ijcnlp-2019}

\end{document}